\documentclass{new_tlp}
\usepackage[utf8]{inputenc}
\usepackage[]{todonotes}
\usepackage{amssymb}
\usepackage[fleqn]{amsmath}
\usepackage[all]{xy}
\usepackage{url}
\usepackage{amsfonts}
\usepackage{mdframed}
\usepackage{url}
\usepackage{hyperref}
\usepackage{soul}
\usepackage{xspace}
\usepackage{multirow}
\usepackage{algorithm}
\usepackage{algpseudocode}
\usepackage[caption = false]{subfig}
\usepackage{float}
\usepackage{tikz}
\usepackage{pdfpages}
\usetikzlibrary{shadows} 
\usetikzlibrary{arrows}
\usetikzlibrary{positioning}
\usetikzlibrary{fit}

\newcommand{\TableProb}{\mathit{TableProb}}

\newcommand{\MAP}{\mathit{MAP}}

\newcommand{\TableMAP}{\mathit{TableMAP}}

\DeclareMathOperator*{\argmax}{arg\,max}

\newcommand{\false}{\mathit{false}}

\newtheorem{definition}{Definition} 
\newtheorem{example}{Example} 

\title{MAP Inference for Probabilistic Logic Programming}
\author[E. Bellodi, M. Alberti, F. Riguzzi, R. Zese] 
{ELENA BELLODI$^1$, MARCO ALBERTI$^2$, FABRIZIO RIGUZZI$^2$, RICCARDO ZESE$^1$\\
$^1$ Dipartimento di Ingegneria -- Universit\`a di Ferrara\\
$^2$ Dipartimento di Matematica e Informatica -- Universit\`a di Ferrara\\
Via Saragat 1, 44122, Ferrara, Italy\\
\email{firstname.surname@unife.it}}

\begin{document}
\maketitle

\begin{abstract}
	
	In Probabilistic Logic Programming (PLP)
	the most commonly studied inference task
	is to compute the marginal probability of a query given a program.
	In this paper, we consider two other important tasks in the PLP setting:
	the Maximum-A-Posteriori (MAP) inference task,
	which determines the most likely values for a subset of the random variables
	given evidence on other variables,
	and the Most Probable Explanation (MPE) task,
	the instance of MAP
	where the query variables are the complement of the evidence variables.
	We present a novel algorithm,
	included in the PITA reasoner,
	which tackles these tasks
	by representing each problem as a Binary Decision Diagram
	and applying a dynamic programming procedure on it.
	We compare our algorithm with the version of ProbLog that admits annotated disjunctions and can perform MAP and MPE inference.  Experiments   on several synthetic datasets show that PITA  outperforms  ProbLog  in many cases.
	This paper is \emph{under consideration} for acceptance in Theory and Practice of Logic Programming.
	
\end{abstract}

\section{Introduction}
\label{sec:intro}

Probabilistic Logic Programming (PLP) \cite{DBLP:conf/ilp/2008p,Rig18-BK} has emerged as one of the most prominent approaches for modeling complex domains containing many uncertain relationships among their entities.
In this field, many languages are equipped with the distribution semantics \cite{DBLP:conf/iclp/Sato95}.
Examples of such languages are
Independent Choice Logic \cite{Poo97-ArtInt-IJ},
PRISM \cite{DBLP:conf/iclp/Sato95},
Logic Programs with Annotated Disjunctions (LPADs) \cite{VenVer04-ICLP04-IC}
and ProbLog \cite{DBLP:conf/ijcai/RaedtKT07}.
All these languages have the same expressive power,
as a theory in one language can be translated into each of the others \cite{DeR-NIPS08}.
LPADs offer a general syntax
as the constructs of all the other languages can be directly encoded in this language.
Under the distribution semantics,  an LPAD defines a probability distribution over a set of normal logic programs called worlds, by associating to each disjunctive clause a
random variable, whose value determines the selection of one of the atoms in the
head.

%

The inference task that has received most attention from the PLP community is computing the marginal probability of a ground query atom $q$ given evidence $e$ on a subset of the other atoms, $P (q | e)$. In the absence of $e$, this is also known as the success probability of a query $P(q)$,
defined as the sum of the probabilities of all the worlds that entail $q$.

Other important inference tasks are the \emph{maximum a posteriori} (MAP) and the \emph{most probable explanation} (MPE) tasks.
In general terms,
given a joint probability distribution over a set of random variables,
values for a subset of the variables (evidence),
and  another disjoint subset of the variables (query),
the MAP problem consists of finding the most probable values for the query variables given the evidence.
The MPE problem is the MAP problem where the set of query variables is the complement of the set of  evidence variables.
In PLP, the MPE problem can be expressed as taking the truth of some atoms as evidence, and finding the world of an LPAD that has the highest probability among those that entail the evidence.
Solving the MAP problem,
given evidence and  a subset of the random variables, consists of finding the assignment to those variables
that maximizes the probability of the assignment given the evidence,
i.e., the sum of the probabilities of the worlds compatible with the assignment and the evidence. 

%


%


The PITA algorithm (for “Probabilistic Inference with Tabling and Answer
subsumption”) \cite{RigSwi10-ICLP10-IC,RigSwi11-ICLP11-IJ,RigSwi13-TPLP-IJ} takes as input an LPAD and computes the probability of success of a query by building Binary Decision Diagrams (BDDs) for every subgoal encountered during the derivation of the query.
In this paper, we present and evaluate experimentally an extension of PITA to perform the MPE and MAP tasks.
We compare PITA to the version of ProbLog presented by \citeN{DBLP:conf/ilp/ShterionovRVKMJ14}, which supports Annotated Disjunctions in the head of clauses (such as LPADs),  allowing to perform the MPE (and MAP) task as well.
ProbLog answers MPE queries by converting each annotated disjunction into a set of probabilistic facts with appropriate probability values and a Prolog rule for each of its head atoms having mutually exclusive bodies, then it generates  the grounding of the resulting program. Then, the program is converted into a Conjunctive Normal Form~(CNF) Boolean formula and knoweldge compilation is applied.  As done by \citeN{DBLP:conf/ilp/ShterionovRVKMJ14} the CNF formula is compiled into a d-DNNF instead of a BDD. d-DNNF are more succinct than BDDs, which means that, given a formula, its d-DNNF version is smaller than its BDD version. 
However, software packages for the manipulation of BDDs are highly optimized and the experiments show that the use of BDDs is sometimes advantageous.
For answering MAP queries ProbLog uses a different strategy resorting to Decision Theoretic ProbLog (DTProbLog) \cite{DBLP:conf/aaai/BroeckTOR10} that exploits Algebraic Decision Diagrams.

We ran experiments on several synthetic datasets;
the results show that PITA performs better than ProbLog on the MAP and MPE tasks in many cases.

The paper is structured as follows:
in Section \ref{sec:background} we summarize the necessary background notions,
in Section \ref{sec:MPE-LPAD} we define the MAP and MPE problems for LPADs, in Section \ref{algorithm} we present their implementation in PITA,
in Section \ref{sec:experimentation} we assess the scalability of our system
and compare it with the same techniques implemented in ProbLog \cite{DBLP:conf/ilp/ShterionovRVKMJ14},
and in Section \ref{sec:conclusions} we conclude the article.

\section{Background}
\label{sec:background}

\subsection{Logic Programs with Annotated Disjunctions}

\newcommand{\pprog}{\ensuremath{\mathcal{P}}\xspace}

LPADs~\cite{VenVer04-ICLP04-IC_iclp} consist of a finite set of annotated
disjunctive clauses $r_i$ of the form
$h_{i1}:\Pi_{i1}; \ldots ; h_{in_i}:\Pi_{in_i} \leftarrow b_{i1},\ldots,b_{im_i}$, where
$b_{i1},\ldots ,b_{im_i}$ are logical
literals that form the \emph{body} of $r_i$, denoted by $body(r_i)$,
while $h_{i1},\ldots h_{in_i}$ are logical atoms and
$\{\Pi_{i1},\ldots,\Pi_{in_i}\}$ are real numbers in the interval
$[0,1]$ such that $\sum_{k=1}^{n_i} \Pi_{ik}\leq 1$.  If $n_i=1$ and
$\Pi_{i1} = 1$ the clause is a non-disjunctive and non-probabilistic clause.  If
$\sum_{k=1}^{n_i} \Pi_{ik}<1$, the head of the annotated disjunctive
clause implicitly contains an extra atom $null$ that does not appear
in the body of any clause and whose annotation is
$1-\sum_{k=1}^{n_i} \Pi_{ik}$.  $ground(\pprog{})$ denotes the
grounding of an LPAD $\pprog{}$. We do not allow function symbols, so
$ground(\pprog{})$ is finite.

\begin{definition}[Variable associated to a clause's grounding]
	To each grounding substitution $\theta_j$ of each clause $r_i$, a discrete random variable $X_{ij}$ is associated, whose range is $0, \ldots, n_i$ and whose probability distribution is given by
	
	$$P(X_{ij} = k) =  \begin{cases}
	\Pi_{ik}                    & \text{if $1 \le k \le n_i$} \\
	1-\sum_{k=1}^{n_i} \Pi_{ik} & \text{if $k = 0$}
	\end{cases} $$
\end{definition}

$X_{ij} = k$ means that the $k$-th head atom, or the $null$ atom if $k = 0$,  is chosen for grounding $\theta_j$ of clause $r_i$.

We now present the distribution semantics for the case in which the
program does not contain function symbols so that its Herbrand base is
finite\footnote{For the distribution semantics with
	function symbols see
	\cite{DBLP:conf/iclp/Sato95,DBLP:journals/jlp/Poole00,RigSwi13-TPLP-IJ,Rig16-IJAR-IJ}.}.

An \emph{atomic choice} is an equation $X_{ij}=k$. A set of atomic choices $\kappa$ is \emph{consistent} if $X_{ij}=k\in\kappa,X_{ij}=m\in \kappa$ implies $k=m$, i.e., only one head is selected for a ground clause.
A  \emph{composite choice} $\kappa$ is a consistent set of atomic choices.
The probability of a composite choice $\kappa$  is
$P(\kappa)=\prod_{X_{ij}=k\in \kappa}P(X_{ij}=k)$.
A \emph{selection} $\sigma$ is a total composite choice (one atomic choice
for every grounding of each probabilistic clause). Let us call $S_T$ the set of all selections.
A selection $\sigma$ identifies a  normal logic program $w_\sigma$ called  a \emph{world}.
The probability of $w_\sigma$ is $P(w_\sigma)=P(\sigma)$.
Since the program does not contain function symbols, the set of worlds  $W_T=\{w_1,\ldots,w_m\}$ is finite and  $P(w)$ is a distribution over worlds:  $\sum_{w\in W_T}P(w)=1$.
The conditional probability of a query $Q$ given a world $w$ can be defined as:
$P(Q|w)=1$ if $Q$ is true in $w$ and 0 otherwise.
We can obtain the probability of the query by marginalizing over the query:
\begin{equation}
P(Q)=\sum_{w}P(Q,w)=\sum_{w}P(Q|w)P(w)=\sum_{w\models Q}P(w)\label{prob}
\end{equation}

\begin{example}
	\label{exnotblue}
	Given the LPAD
	{\small   \begin{verbatim}
		red(b1):0.6; green(b1):0.3; blue(b1):0.1 :- pick(b1).
		pick(b1):0.6; no_pick(b1):0.4.
		ev:- \+ blue(b1).
		\end{verbatim}}
	the query \verb|ev| is true in five worlds so its probability is
	$P(\verb|ev|)=0.6\cdot 0.6+0.6\cdot 0.3+ 0.4\cdot 0.6+0.4\cdot 0.3+0.4\cdot 0.1=0.94$.
\end{example}

A composite choice $\kappa$ \textit{identifies} a set $\omega_\kappa$ that contains all the worlds associated with a selection that is a superset of $\kappa$: i.e.,
$\omega_\kappa=\{w_\sigma|\sigma \in S_T, \sigma \supseteq \kappa\}$.
We define the set of worlds \textit{identified} by a set of composite choices $K$ as
$\omega_K=\bigcup_{\kappa\in K}\omega_\kappa$.
Given a ground literal $Q$, a composite choice $\kappa$ is an \emph{explanation} for $Q$ if $Q$ is true in every world of $\omega_\kappa$.
A set of composite choices $K$ is \emph{covering} with respect to $Q$ if every world $w_\sigma$ in which $Q$ is true is such that $w_\sigma\in\omega_K$.
Given a covering set of explanations for a query, we can obtain a Boolean formula $f(\mathbf{X})$ in Disjunctive Normal Form (DNF) where: (1)~each atomic choice yields an equation $X_{ij}=k$, (2) we replace an explanation with the conjunction of the equations of its atomic choices and the set of explanations with the disjunction of the formulas for all explanations.
If we consider a world as the specification of a truth value for each equation $X_{ij}=k$, the formula    evaluates to true exactly on the worlds where the query is true  \cite{DBLP:journals/jlp/Poole00}.
Since the disjuncts in the formula are not necessarily mutually exclusive, the probability of the query can not be computed by a summation as in Formula (\ref{prob}). The problem of computing the probability of a Boolean formula in DNF, known as \textit{disjoint sum}, is \#P-complete
\cite{valiant1979complexity}. One of the most effective ways  of solving
the problem makes use of Decision Diagrams.

\subsection{Binary  Decision Diagrams}
\label{bddsec}
We can apply \textit{knowledge compilation} \cite{DBLP:journals/jair/DarwicheM02} to the Boolean formula $f(\mathbf{X})$ in order to translate it into a ``target language''
that allows the computation of its probability in polynomial time. We can use Decision Diagrams (DD) as a target language. A DD has one level for each variable and two leaves, one associated with the 1 Boolean function and the other with the 0 Boolean function.
Each variable node has as many children as its values. A DD can be used to compute the value of a Boolean function given the values of the variables by starting at the root and following the path according to the variable values until a leaf is reached. The label of the leaf is the value of the Boolean function.
Most packages for the manipulation of  DDs are however restricted to work on Binary Decision Diagrams (BDD), i.e., decision diagrams where all the variables are Boolean.
These packages offer Boolean operators among BDDs and apply simplification rules to the results of operations in order to reduce as much as possible the size of the  diagram, producing a  reduced BDD.

\begin{figure}[ht]
	\centering
	\begin{minipage}[b]{0.63\linewidth}
		\begin{algorithm}[H]\begin{scriptsize}
				\caption{Function Prob: computation of the probability of a BDD.} \label{alg:prob}
				\begin{algorithmic}[1]
					\Function{Prob}{$node$}
					\If{$node$ is a terminal}
					\State  \Return $1$
					\Else
					\If{$\TableProb(node.pointer)\neq null$}
					\State\Return $\TableProb(node)$
					\Else
					\State $p0\gets$\Call{Prob}{$child_0(node)$}
					\State $p1\gets$\Call{Prob}{$child_1(node)$}
					\If{$child_0(node).comp$}
					\State $p0\gets (1-p0)$
					\EndIf
					\State Let $\pi$ be the probability of being true of $var(node)$
					\State $Res\gets p1\cdot \pi+p0\cdot (1-\pi)$\label{aggreg}
					\State Add $node.pointer\rightarrow Res$ to $\TableProb$
					\State\Return $Res$
					\EndIf
					\EndIf
					\EndFunction
				\end{algorithmic}
			\end{scriptsize}
		\end{algorithm}
	\end{minipage}
	\quad
	\begin{minipage}[b]{0.33\linewidth}
		\includegraphics[width=0.95\textwidth]{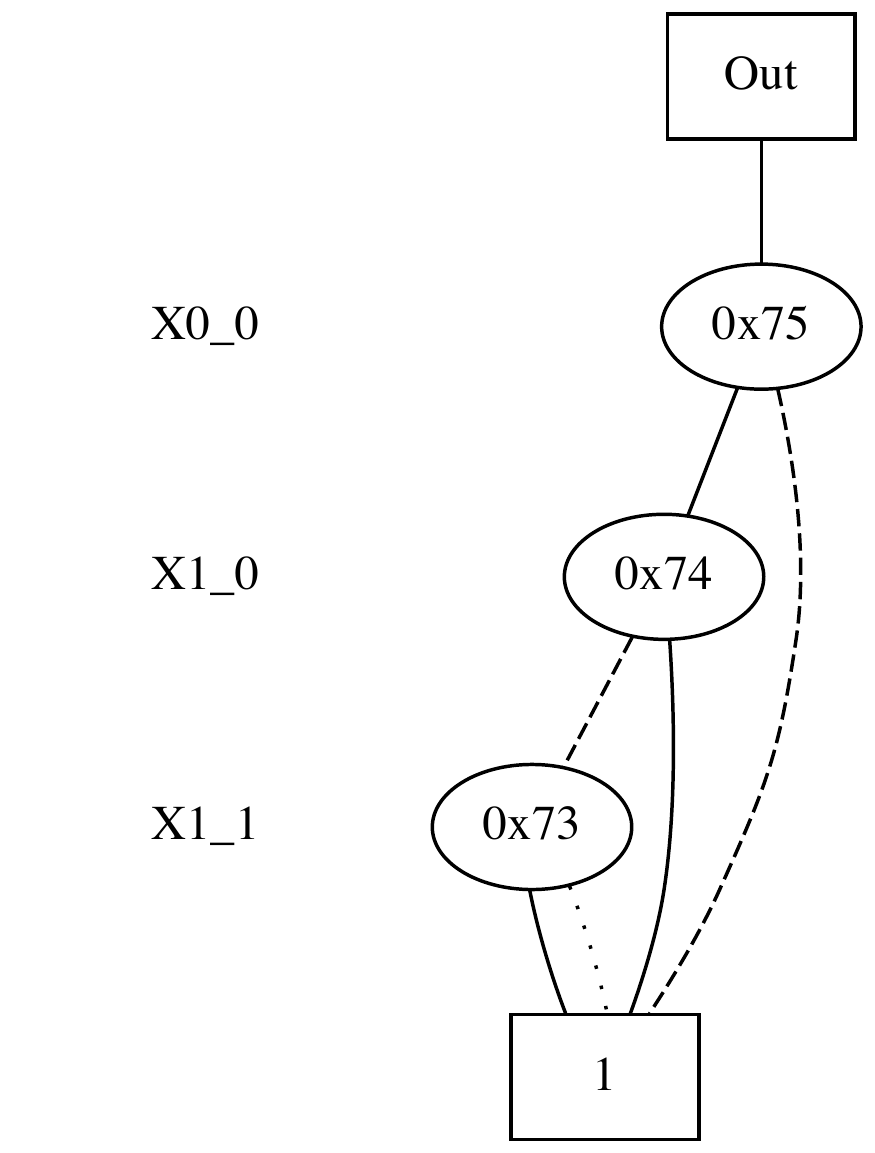}
		\caption{BDD for Example \ref{exnotblue}.} \label{bddnotblue}
	\end{minipage}
\end{figure}

A node $n$ in a BDD has two children: the 1\--child and the 0\--child.
To work with a BDD package we must represent multi\--valued variables by means of binary variables. We use the following encoding, called \emph{order encoding}: for a multi\--valued variable $X_{ij}$, corresponding to a ground clause $C_{i}\theta_j$, having $n_i$ values, we use $n_i - 1$ Boolean variables $X_{ij1},\ldots,X_{ijn_i-1}$ and we represent the equation $X_{ij}=k$ for $k=1,\ldots n_i-1$ by means of the conjunction  $\overline{X_{ij1}}\wedge\ldots \wedge \overline{X_{ijk-1}}\wedge X_{ijk}$, and the equation $X_{ij}=n_i$ by means of the conjunction  $\overline{X_{ij1}}\wedge\ldots \wedge\overline{X_{ijn_i}}$. Note that $\lceil \log_2  n_i\rceil$ binary variables would be sufficient to represent an $n_i$-valued variable, but the encoding that we use allows for faster BDD processing.
A parameter $\pi_{ik}$ is associated with
each Boolean variable $X_{ijk}$.
The parameters are obtained from those of multi\--valued variables in this way:
$\pi_{i1}=\Pi_{i1}$,
$\ldots$,
$\pi_{ik}=\frac{\Pi_{ik}}{\prod_{j=1}^{k-1}(1-\pi_{ij})}$,
up to $k=n_i-1$.
%
In order to manage BDD we exploit the CUDD (Colorado University Decision Diagram)\footnote{\url{https://github.com/ivmai/cudd}}
library, a library written in C that provides functions to manipulate different types of Decision Diagrams. In CUDD, BDD nodes are described by two fields: $pointer$, a pointer to the node, and $comp$, a Boolean indicating whether the node is complemented. In fact  three types of edges are admitted: an edge to a 1-child, an edge to a 0-child and a complemented edge to a 0-child, meaning that the function encoded
by the child must be complemented.
Moreover, the root node can be complemented. For these types of BDD, only the 1 leaf is needed. Once a BDD for a query has been built, it is possible to compute the probability of the query using a dynamic programming algorithm \cite{raedt07problog}, which is shown in Algorithm \ref{alg:prob}.


The BDD for the query \verb|ev| from Example \ref{exnotblue} is shown in Figure \ref{bddnotblue}, where edges going to the 1-child are solid, edges going to the 0-child are dashed and complemented edges going to  the 0-child are dotted. Variables X1\_0 and X1\_1 encode the first rule and variable X0\_0 the second rule. Node labels are just identifiers.

\section{MAP and MPE Inference for LPADs Programs}
\label{sec:MPE-LPAD}

\begin{definition}[MAP Problem]
	Given an LPAD $\pprog{}$, a conjunction of ground atoms $e$, the \emph{evidence}, and a set of random variables $\mathbf{X}$ (query random variables), associated to some  ground rules of $\pprog{}$,
	the MAP problem is to find an assignment $\mathbf{x}$ of  values to $\mathbf{X}$ such that $P(\mathbf{x}|e)$ is maximized, i.e., solve
	$$\argmax_{\mathbf{x}}P(\mathbf{x}|e)$$
	The MPE problem is a MAP problem where $\mathbf{X}$ includes all the random variables associated with all  ground clauses of $\pprog{}$.
\end{definition}
In the following, we indicate  the query random variables  in the program by prepending the functor \verb|map_query| to the rules.

%
\citeN{DBLP:conf/ilp/ShterionovRVKMJ14} showed that the encoding presented in Section \ref{bddsec} using $n_i-1$ Boolean variables for a clause with $n_i$ heads does not work, as configurations of the variables exist that do not correspond to any value for the
rule random variable. 
The problem is that the order encoding is redundant and a value for the random variable associated with a rule may be encoded by multiple tuples of values of the Boolean variables besides the intended one. One of those unintended encodings may get chosen because it has a higher probability but this does not reflect on the correct choice of the multivalued variable.
\citeN{DBLP:conf/ilp/ShterionovRVKMJ14} proposed a different encoding, where $n_i$ Boolean variables $X_{ijk}$ for a clause with $n_i$ heads are used and constraints are imposed,
namely that one and only one $X_{ijk}$ must be true. This is achieved by building the constraint formula
$$(\bigvee_{k=1}^{n_i}X_{ijk})\wedge\bigwedge_{k=1}^{n_i}\bigwedge_{m=k+1}^{n_i}(\neg X_{ijk}\vee \neg X_{ijm})\label{const}$$
for each multi-valued variable $X_{ij}$, translating it into a BDD and conjoining it with the  BDD built for the query.

\begin{figure}[ht]
	\centering
	\begin{minipage}[b]{0.45\linewidth}
		\centering
		\includegraphics[width=0.7\textwidth]{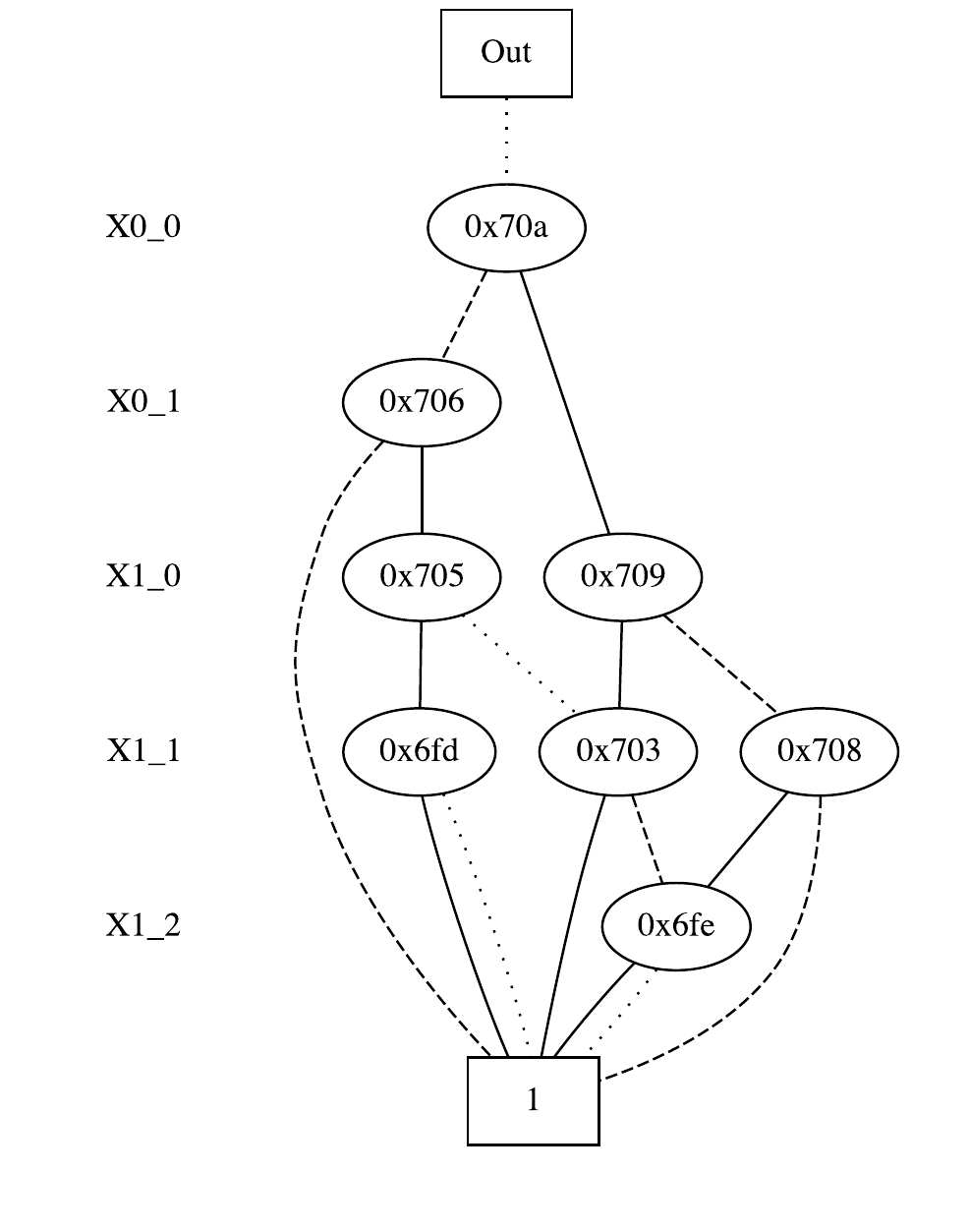}
		\caption{BDD for the MPE problem of Example \ref{exnotblue_MPE}.}
		\label{bddes2}
	\end{minipage}
	\quad
	\begin{minipage}[b]{0.45\linewidth}
		\centering
		\includegraphics[width=0.55\textwidth]{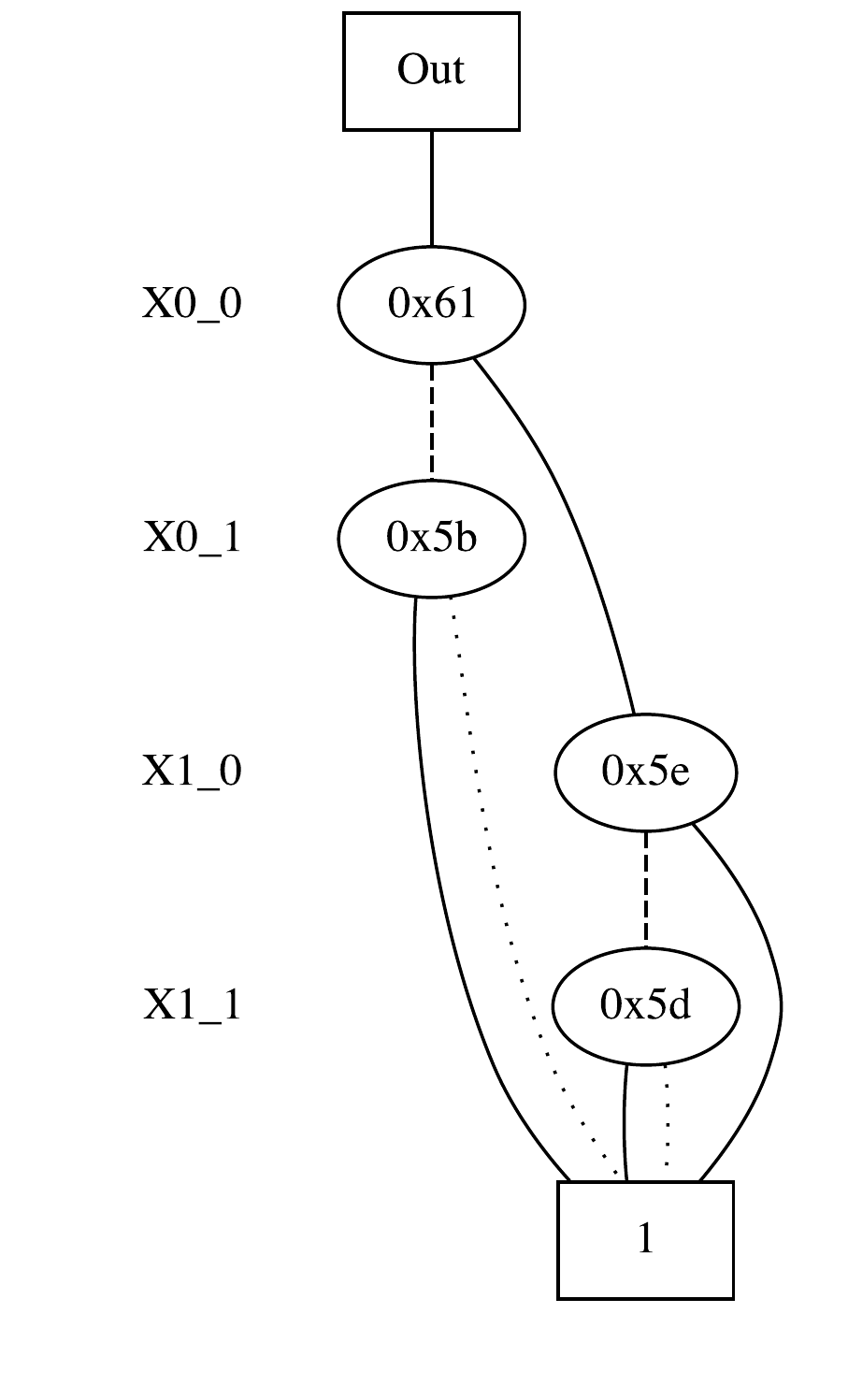}
		\caption{BDD for the MAP problem of Example \ref{exnotblueMAP}.}
		\label{bddmapes2_map}
	\end{minipage}
\end{figure}

\begin{example}
	\label{exnotblue_MPE}
	Given the program of Example \ref{exnotblue}
	{\small   \begin{verbatim}
		map_query red(b1):0.6; green(b1):0.3; blue(b1):0.1 :- pick(b1).
		map_query pick(b1):0.6; no_pick(b1):0.4.
		ev:- \+ blue(b1).
		\end{verbatim}}
	where all the random variables are query, evidence \verb|ev| has the MPE assignment $\mathbf{x}$:
	{\small   \begin{verbatim}
		[rule(1, pick(b1), [pick(b1):0.6, no_pick(b1):0.4], true), 
		rule(0, red(b1), [red(b1):0.6, green(b1):0.3, blue(b1):0.1], pick(b1))],
		\end{verbatim}}
	where predicate \verb|rule/4| specifies  clause number (zero-based),  selected head,  clause head,  clause body, in that order.
	For this assignment, $P(\mathbf{x}|ev)= 0.36$,  meaning that the most probable explanation $\mathbf{x}$ has a probability of 0.36.
	The corresponding BDD  is shown in Figure \ref{bddes2},  where variables X0\_k are associated with the second clause and X1\_k with the first  clause.
\end{example}


\begin{example}
	\label{exnotblueMAP}
	Given the program
	{\small   \begin{verbatim}
		red(b1):0.6; green(b1):0.3; blue(b1):0.1 :- pick(b1).
		map_query pick(b1):0.6; no_pick(b1):0.4.
		ev:- \+ blue(b1).
		\end{verbatim}}
	The evidence  \verb|ev| has the MAP assignment:
	{\small   \begin{verbatim}
		[rule(1, pick(b1), [pick(b1):0.6, no_pick(b1):0.4], true)].
		\end{verbatim}}
	For this assignment, $P(\mathbf{x}|ev)= 0.54$.
	The corresponding BDD is shown in Figure \ref{bddmapes2_map}, where variables X0\_k are associated to the second rule  and X1\_k to the first rule.
\end{example}

\begin{example}
	Consider the following LPAD:
	

	{\small   \begin{verbatim} 
		map_query disease:0.05.
		map_query malfunction:0.05.
		positive :- malfunction.
		map_query positive:0.999 :- disease.
		map_query positive:0.0001 :- \+(malfunction), \+(disease).
		\end{verbatim}}
	The LPAD models the diagnosis of a disease by means of a lab test. The disease probability is $0.05$, and, in case of disease, the test result will be positive with probability $0.999$. However, there is a 5\% chance of an equipment malfunction; in this case, the test will always be positive. Additionally, even in absence of disease or malfunction, the test result will be positive with probability $0.0001$.
	The LPAD has 16 worlds, each corresponding to selecting, or not, the head of each  annotated disjunctive clause.
	
	\noindent Let us suppose for the test result to be positive:
	is the patient ill?
	Given evidence \verb+ev = positive+, the MPE assignment is
	
	{\small   \begin{verbatim}    
		[rule(1, '', [malfunction:0.05, '' :0.95], true),
		rule(0, disease, [disease:0.05, '' :0.95], true),
		rule(2, positive, [positive:0.999, '' :0.001], disease),
		rule(3, '', [positive:0.0001, '' :0.9999], (\+malfunction,\+disease))]
		\end{verbatim}}
	\noindent where \verb|''| indicates the $null$ head. The most probable world is the one where
	an actual disease caused the positive result, and its probability is  $P(\mathbf{x}|e) = 0.04702$.
	
	Likewise, if we perform a MAP inference taking only the choice of the first clause as query variable,
	the result is \verb+[rule(0, disease, [disease:0.05, '' : 0.95], true)]+,
	so the patient is ill. However, if we take the choices for the first two clauses as query variables,
	i.e., if we look for the most likely combination of \verb+disease+ and \verb+malfunction+ given \verb+positive+,
	the MAP task produces 
	
	{\small   \begin{verbatim}
		[rule(1, malfunction, [malfunction:0.05, '' : 0.95], true),
		rule(0, '', [disease:0.05, '' : 0.95], true)]
		\end{verbatim}}
	\noindent meaning that the patient is not ill
	and the positive test is explained by an equipment malfunction.
	This examples shows that the value assigned to a query variable in a MAP task
	can be affected by the presence of other variables in the set of query variables;
	in particular, MPE and MAP inference over $\mathbf{X}$
	may assign different values to the same variable given the same evidence.
	
	\label{ex:disease}
\end{example}

\section{Integration of MAP and MPE Inference into the PITA System}
\label{algorithm}

PITA (Probabilistic Inference with Tabling and Answer subsumption)
\cite{RigSwi10-ICLP10-IC,RigSwi13-TPLP-IJ} computes the probability of a query from a probabilistic program
in the form of an LPAD by first transforming the LPAD into a normal
program containing calls for manipulating BDDs.  The
idea is to add an extra argument to each subgoal to store a BDD encoding the explanations for the answers of the subgoal.  The values of the subgoals' extra argument
are combined using a set of general library functions:
\begin{itemize}
	\item \texttt{init, end}: initialize and terminate the  data structures  for manipulating BDDs;
	\item \texttt{zero(-D),one(-D)}: return the BDD \texttt{D} representing the Boolean constants 0, 1;
	\item \texttt{and(+D1,+D2,-DO), or(+D1,+D2,-DO), not(+D1,-DO)}: Boolean operations among BDDs;
	\item \texttt{equality(+Var,+Value,-D)}: \texttt{D} is the BDD representing \texttt{Var=Value}, i.e.  the multi\--valued random variable \verb|Var| is assigned \verb|Value|;
	\item \texttt{ret\_prob(+D,-P)}: returns the probability \verb|P| of the BDD \texttt{D}.
\end{itemize}
These functions are implemented in C as an interface to the CUDD library for manipulating BDDs. A BDD is represented in Prolog as an integer that is a pointer in memory to its root node.

\begin{figure}
	\subfloat[Node for variable $a$ in a BDD.\label{bddportion}]{
		\centering
		\begin{tikzpicture}
		[every node/.style={font=\tiny},x=1cm,y=0.7cm,
		zeroarrow/.style = {dashed},
		onearrow/.style = {solid},
		c/.style = {circle,draw,solid},
		r/.style = {rectangle,draw,solid,fill=black!10!white}]
		
		\node[c] (root) at (0,0) {$a$};
		\node[c] (a) at (-1,-1) {$\alpha$};
		\node[c] (b) at (1,-1) {$\beta$};
		\path[zeroarrow] (root) edge[-latex] (b);
		\path[onearrow] (root) edge[-latex] (a);
		\end{tikzpicture}
	}
	\hspace{1cm}
	\subfloat[d-DNNF equivalent portion.\label{ddnnf}]{
		\centering
		\begin{tikzpicture}
		[every node/.style={font=\tiny},x=1cm,y=0.7cm,
		zeroarrow/.style = {dashed},
		onearrow/.style = {solid},
		c/.style = {circle,draw,solid},
		r/.style = {rectangle,draw,solid,fill=black!10!white}]
		
		\node[c] (root) at (0,0) {$\vee$};
		\node[c] (wedge1) at (-1,-1) {$\wedge$};
		\node[c] (wedge2) at (1,-1) {$\wedge$};
		
		\node[r] (X) at (-1.5,-2) {$a$};
		\node[c] (a) at (-0.5,-2) {$\alpha'$};
		\node[r] (nX) at (0.5,-2) {$\neg a$};
		\node[c] (b) at (1.5,-2) {$ \beta'$};
		
		\path[onearrow] (root) edge[-latex] (wedge1);
		\path[onearrow] (root) edge[-latex] (wedge2);
		\path[onearrow] (wedge1) edge[-latex] (X);
		\path[onearrow] (wedge1) edge[-latex] (a);
		\path[onearrow] (wedge2) edge[-latex] (nX);
		\path[onearrow] (wedge2) edge[-latex] (b);
		\end{tikzpicture}
	}
	\hspace{1cm}
	\subfloat[Arithmetic circuit.\label{ac}]{
		\centering
		\begin{tikzpicture}
		[every node/.style={font=\tiny},x=1cm,y=0.7cm,
		zeroarrow/.style = {dashed},
		onearrow/.style = {solid},
		c/.style = {circle,draw,solid},
		r/.style = {rectangle,draw,solid,fill=black!10!white}]
		
		\node[c] (root) at (0,0) {$\max$};
		\node[c] (wedge1) at (-1,-1) {$\times$};
		\node[c] (wedge2) at (1,-1) {$\times$};
		
		\node[r] (X) at (-1.5,-2) {$a$};
		\node[c] (a) at (-0.5,-2) {$\alpha$};
		\node[r] (nX) at (0.5,-2) {$\neg a$};
		\node[c] (b) at (1.5,-2) {$ \beta$};
		
		\path[onearrow] (root) edge[-latex] (wedge1);
		\path[onearrow] (root) edge[-latex] (wedge2);
		\path[onearrow] (wedge1) edge[-latex] (X);
		\path[onearrow] (wedge1) edge[-latex] (a);
		\path[onearrow] (wedge2) edge[-latex] (nX);
		\path[onearrow] (wedge2) edge[-latex] (b);
		\end{tikzpicture}
	}
	\caption{Translation from BDDs to d-DNNF.\label{bdd2ddnnf}}
\end{figure}
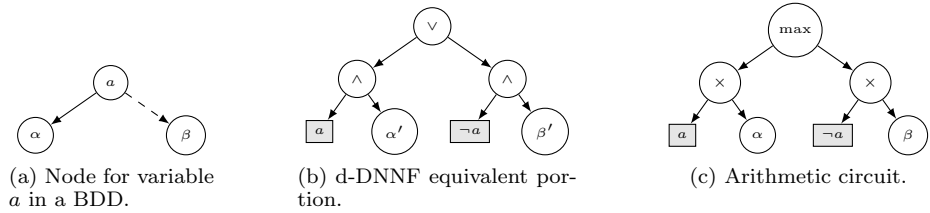

Let us first consider the MPE task. PITA solves it using the dynamic programming algorithm proposed by  \citeN[Section 12.3.2]{DBLP:conf/ecai/Darwiche04} for computing MPE over d-DNNFs, which define a propositional language that generalizes BDDs.
In fact, a BDD can be seen as a d-DNNF by using the translation shown in Figure \ref{bdd2ddnnf}: a BDD node (Figure \ref{bddportion}) for variable $a$ with children $\alpha$ and $\beta$ is translated into the d-DNNF portion shown in Figure~\ref{ddnnf}, where $\alpha'$ and $\beta'$ are the translations of the BDD $\alpha$ and $\beta$ respectively.
The algorithm proposed by  \citeN{DBLP:conf/ecai/Darwiche04} computes the probability of the MPE by replacing $\wedge$-nodes with product nodes and $\vee$-nodes with $\max$-nodes: the result is an arithmetic circuit (Figure \ref{ac}) that, when evaluated bottom-up, gives the probability of the MPE   and can be used to identify the MPE assignment. The equivalent algorithm operating on BDDs - Function \textsc{MAPInt} in Algorithm \ref{map} -  modifies Algorithm \ref{alg:prob} and returns both a probability and a set of assignments to random variables. At each node, instead of computing $Res\gets p1\cdot \pi+p0\cdot (1-\pi)$ as in Algorithm~\ref{alg:prob}  line \ref{aggreg}, it returns the assignment of the children having the maximum probability. This is computed in lines  \ref{check}-\ref{endcheck} in  Algorithm~\ref{map}.
In MPE there are no non-query variables, so the test in line \ref{checkquery} succeeds only for the BDD leaf.
\textsc{MAPInt} in practice computes the probability of paths from the root to the 1 leaf and returns the probability and the assignment corresponding to the most probable path.
In a MAP task, i.e., when we have non-query variables, function \textsc{MAPInt} cannot be used because when a node for a non-query variable is reached, it must be summed out instead of maximized out, and maximization and summation operations are not commutative. However, if its children are nodes for query variables, which of the two assignments for the children should be propagated towards the root? If query variables are mixed with non-query variables in the BDD variable ordering, function \textsc{MAPInt} does not work.
In case that the non-query variables appear last in the ordering, when \textsc{MAPInt} reaches a node for a non-query variable, it can sum out all non-query variables using function \textsc{Prob} from Algorithm \ref{alg:prob}. This assigns a probability to the node that can be used by \textsc{MAPInt} to identify the most probable path from the root.
So PITA solves MAP  by reordering  variables in the BDD, putting first the query variables.
With CUDD  we can either create BDDs from scratch with a given variable order or  modify  BDDs according to a new variable order. 
Changing the position of a variable is made by successive swapping of adjacent variables~\cite{DBLP:journals/sttt/Somenzi01}: the swap can be performed in a time proportional to the number of nodes associated with the two swapped variables. 
Changing the order of two adjacent variables does not affect the other levels of the BDD, so 
changes can be applied directly to the current BDD saving memory. To further reduce the cost of the swapping, the CUDD library keeps in memory an interaction matrix specifying which variables directly interact with others. This matrix is updated only when a new variable is inserted into the BDD, is symmetric and can be stored by using a single bit for each pair, making it very small. Moreover, the cost of building it is negligible compared to the cost of manipulating the BDD without checking it.
Jiang et al. empirically demonstrated that changing the order of variables by means of sequential swapping is usually much more time efficient than rebuilding the BDD following a fixed variable order~\cite{jiang2017variable}.

PITA differs from ProbLog in both tasks. For MPE inference, ProbLog applies the algorithm of (Darwiche 2014) to d-DNNF. For MAP, ProbLog uses DTProbLog, an algorithm for maximizing an utility function by making decisions. In DTProbLog utility values are assigned to some ground literals, some ground atoms are probabilistic and some are decision.
The aim is to find an assignment to decision variables that maximizes  utility, given by the sum of the utility for the literals that are made true by the decisions. DTProbLog uses Algebraic Decision Diagrams (ADDs) as a target compilation language. ADDs are BDDs where  leaves are associated with real numbers instead of  Boolean values.  ADDs built by DTProbLog contain only decision variables, probabilistic variables are compiled away. We differ from DTProbLog because we do not compile away non-query variables but we simply rearrange the BDD. As shown by the experiments, this is sometimes advantageous.

\begin{algorithm}[ht]\begin{scriptsize}
		\caption{ Function MAP: computation of the maximum a posterior state of a set of query variables and of its probability} \label{map}
		\begin{algorithmic}[1]
			\Function{MAP}{$root$}
			
			\ForAll{query variables $var$}
			\State $AtLeastOne\gets BDD\_Zero$
			\State $AtMostOne\gets BDD\_One$
			\For{$i\gets1$ to $values(var)$}
			\State $AtLeastOne\gets BDD\_Or(AtLeastOne,bVar(var,i))$
			\For{$j\gets i+1$ to $values(var)$}
			\State $NotBoth\gets BDD\_Not(BDD\_And(bVar(var,i),bVar(var,j)))$
			\State $AtMostOne\gets BDD\_And(AtMostOne,NotBoth)$
			\EndFor
			\EndFor
			\State $const\gets BDD\_And(AtLeastOne,AtMostOne)$
			\State $root\gets BDD\_And(root,const)$
			\EndFor
			\State Reorder BDD  $root$ so that  variables associated to query variables come first in the order
			\State Let $root'$ be the new root
			\State $\TableMAP\gets \emptyset$
			\State $\TableProb\gets \emptyset$
			\State $(\mathit{Prob},\MAP)\gets$\Call{MAPInt}{$root,\false$} \Comment{MAPBV: map assignment for Boolean random variables}
			\State \Return $(\mathit{Prob},\MAP)$
			\EndFunction
			\Function{MAPInt}{$node,comp$} \Comment{Internal function implementing the dynamic programming algorithm}
			\State $comp\gets node.comp \oplus comp$
			\If{$var(node)$ is not associated to a query var}\label{checkquery}
			\State $p\gets $\Call{Prob}{$node$} 	\Comment{Algorithm \ref{alg:prob}}
			\If{$comp$}
			\State\Return $(1-p,[])$
			\Else
			\State\Return $(p,[])$
			\EndIf
			\Else
			\If{$\TableMAP(node.pointer)\neq null$}
			\State\Return $\TableMAP(node.pointer)$
			\Else
			\State $(p0,\MAP0)\gets$\Call{MAPInt}{$child_0(node),comp$}
			\State $(p1,\MAP1)\gets$\Call{MAPInt}{$child_1(node),comp$}
			\State Let $\pi$ be the probability of being true of the variable at level $level$
			\State $p1\gets p1\cdot  \pi$
			\If{$p1>p0$}\label{check}
			\State $Res\gets (p1,[var(node)=1|\MAP1])$
			\Else
			\State $Res\gets (p0,\MAP0)$
			\EndIf\label{endcheck}
			\State Add $(node.pointer)\rightarrow Res$ to $\TableMAP$
			\State\Return $Res$
			\EndIf
			\EndIf
			\EndFunction
		\end{algorithmic}
	\end{scriptsize}
\end{algorithm}

\section{Experimental Results}
\label{sec:experimentation}

Experiments aim at analyzing how PITA  scales when doing MAP and MPE inference w.r.t. the data size,
and at comparing their performance with the same tasks performed by ProbLog2.1 \cite{DBLP:journals/corr/abs-1304-6810} in terms of inference time. 

Experiments were performed on GNU/Linux machines with Intel Xeon E5-2697 v4 (Broadwell) at 2.30 GHz and 128 GB of RAM available and were set to a maximum execution time   of 24h.
Four artificially generated datasets were used: \textit{growing head (gh)}, \textit{growing negated body (gnb)}, \textit{blood} \cite{DBLP:conf/ilp/ShterionovRVKMJ14},  and \textit{probabilistic graphs}. \textit{Growing head} is a set of  15 programs with annotated disjunctions with an increasing number of head atoms; \textit{growing negated body} is a set of 50 programs with an increasing number of negated body atoms; \textit{blood} is a set of 100 programs regarding the inheritance of blood type with an  increasing number of ancestors (mother+father for each person); \textit{probabilistic graphs} is a set of $N \times M$ programs, where $N=\{50,100,150,200,250,300,400,450,500\}$ is the number of nodes of the graphs and $M=10$ is the number of different probabilistic edge configurations for each graph size. The graphs have been randomly generated  according to the Barab\'asi-Albert model \cite{Barabasi99} with parameters $m_o=m=2$.
These benchmarks can be found at \url{http://ml.unife.it/material/}.
In the following, results are commented separately for MAP and MPE inference.

\subsection{MPE Results}
\label{mpe_res}
\begin{figure}[ht]
	\centering
	\begin{minipage}[b]{0.48\linewidth}
		\includegraphics[width=.9\textwidth]{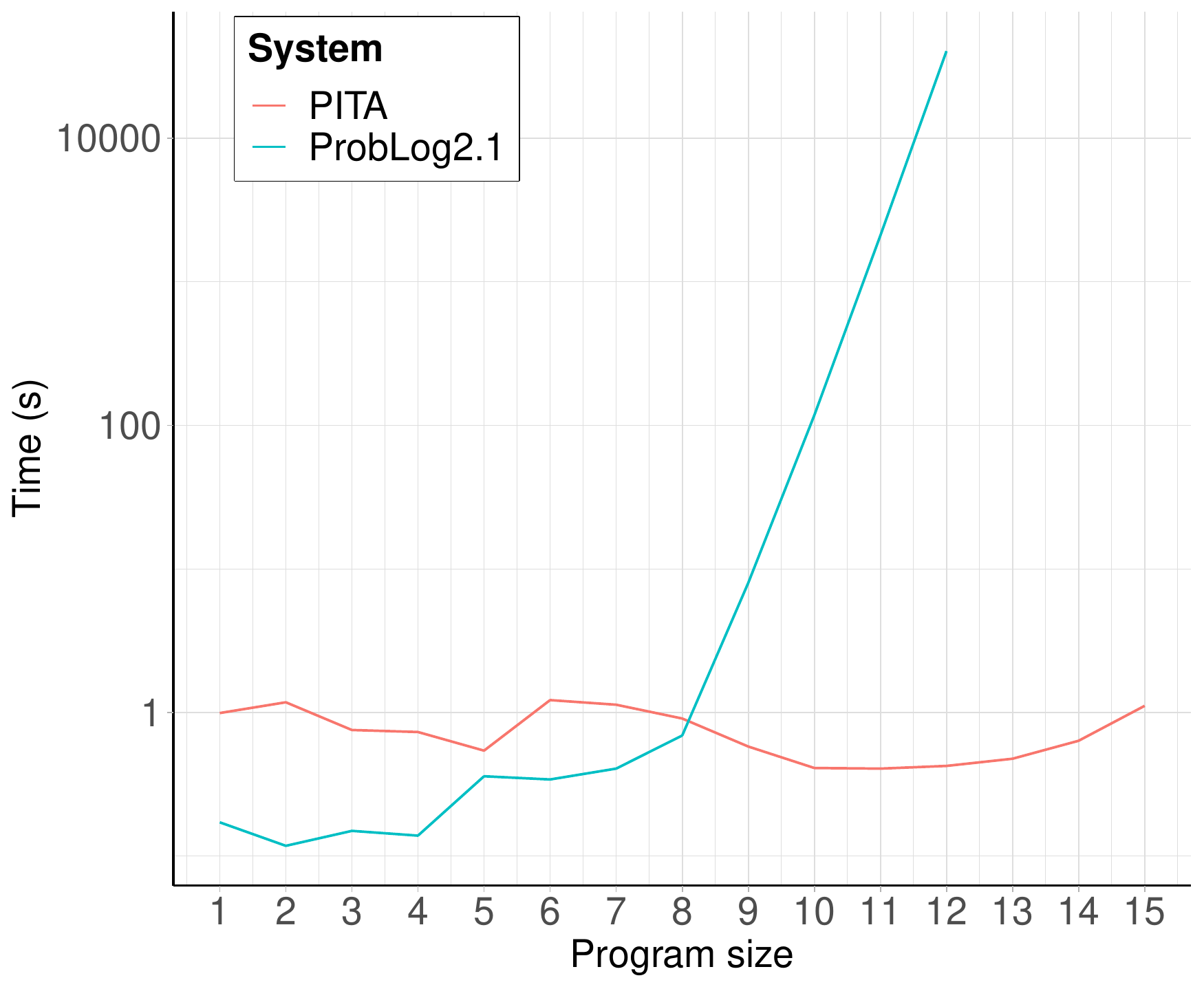}
		\caption{MPE results on the \textit{growing head} dataset (log scale on Y axis).}
		\label{gh}
	\end{minipage}
	\quad
	\begin{minipage}[b]{0.48\linewidth}
		\includegraphics[width=0.9\textwidth]{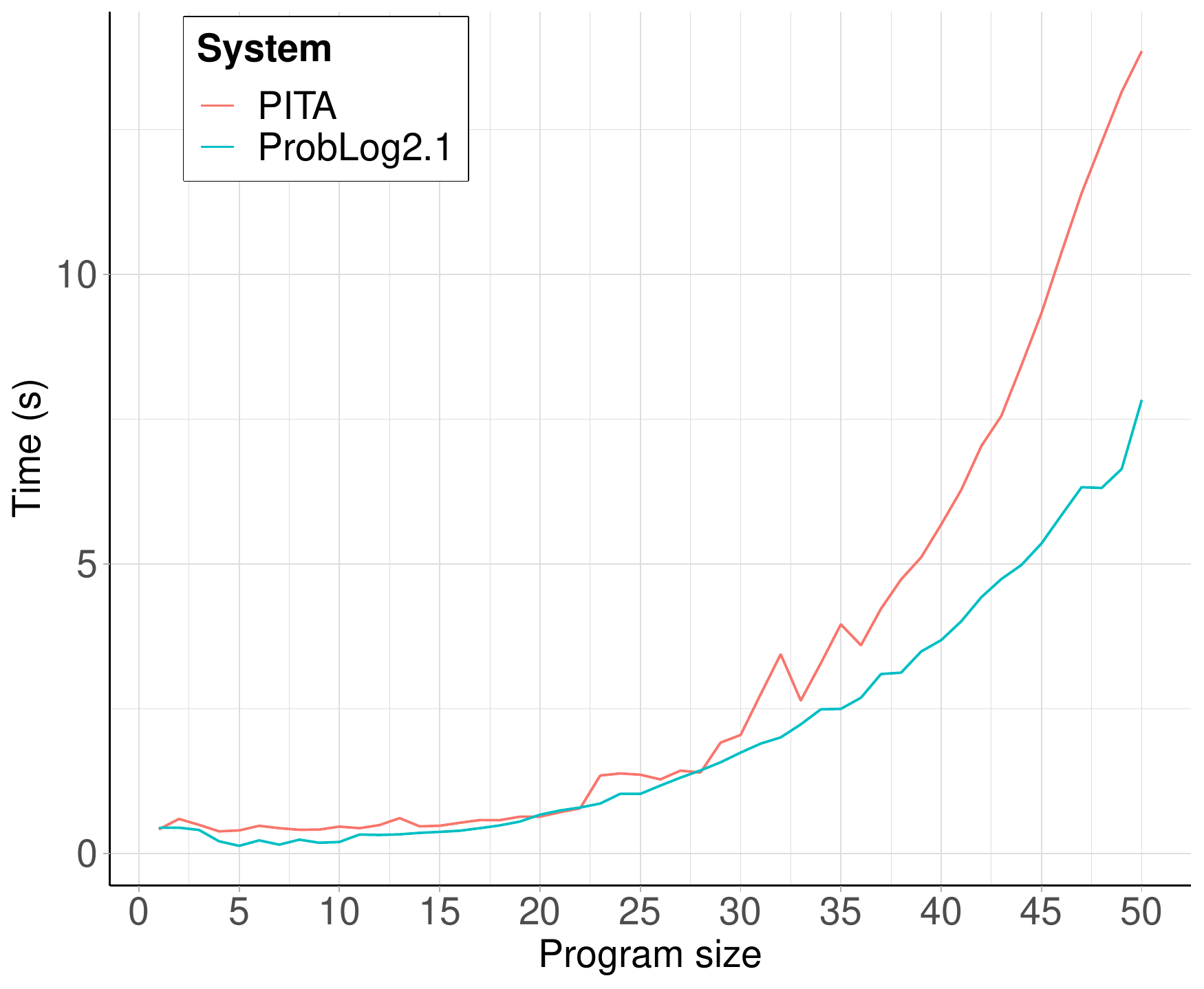}
		\caption{MPE results on the \textit{growing negated body} dataset.}
		\label{gnb}
	\end{minipage}
\end{figure}

\begin{figure}[ht]
	\centering
	\begin{minipage}[t]{0.48\linewidth}
		\includegraphics[width=0.9\textwidth]{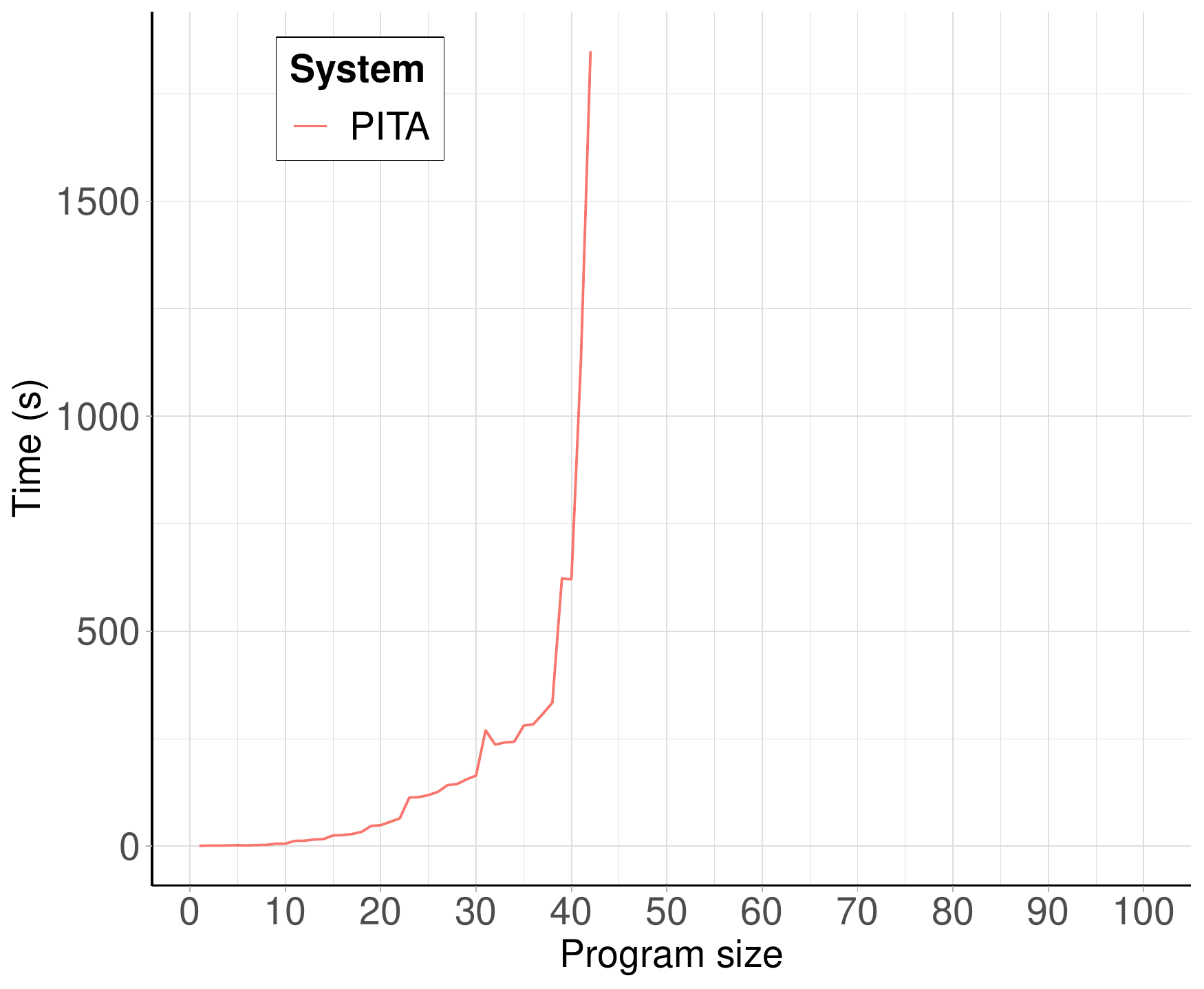}
		\caption{MPE results on the \textit{blood} dataset.}
		\label{blood}
	\end{minipage}
	\quad
	\begin{minipage}[t]{0.48\linewidth}
		\includegraphics[width=0.9\textwidth]{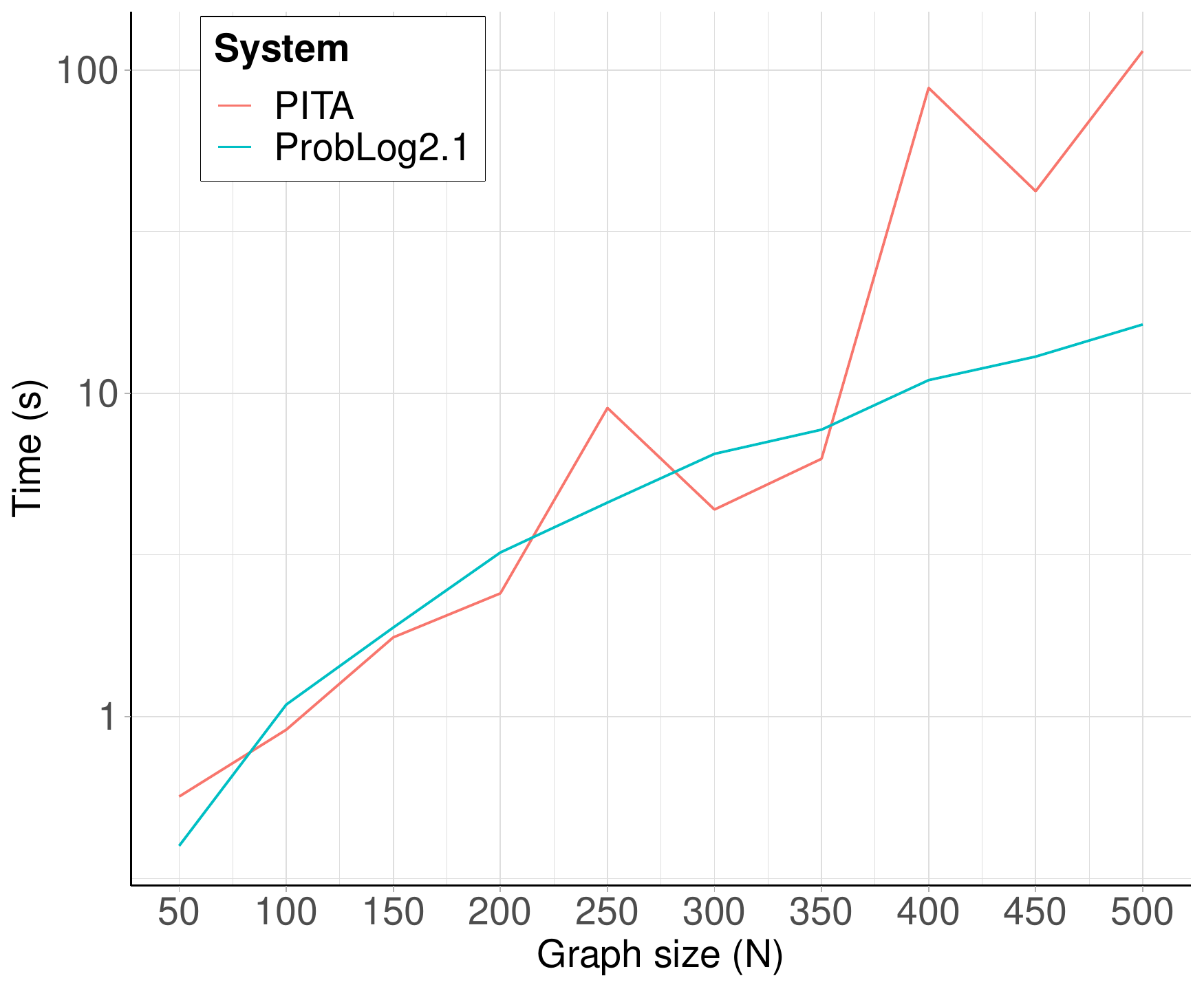}
		\caption{MPE results on the \textit{probabilistic graphs} dataset (log scale on Y axis).}
		\label{graphs}
	\end{minipage}
\end{figure}

For these experiments we ran PITA  and ProbLog 2.1 on all datasets, except for \textit{blood} on which only PITA could be applied due to Problog2.1 execution timing out.
%
ProbLog2.1 was run with the command \texttt{problog-cli.py mpe} \textit{program.pl}. This system requires  to specify  evidence in \textit{program.pl} with the  \verb|evidence/1| fact.
For \textit{gh} and \textit{gnb}, evidence corresponds to \verb|a0|, for \textit{blood} to \verb|bloodtype(p,a)|, for \textit{probabilistic graphs} to \verb|path(0,N-1)| (e.g. \verb|path(0,49)| when $N=50$).
Inference times are compared in Figures \ref{gh}, \ref{gnb}, \ref{blood}, \ref{graphs}; for \textit{probabilistic graphs}  the average time over the 10 configurations for each $N$ was computed. PITA outperforms ProbLog on \textit{gh} and \textit{blood}, where the latter times out starting from program size 13 or from the beginning, respectively; on \textit{gnb}  and \textit{probabilistic graphs} the systems are comparable for small program sizes, then PITA is slower. This shows that, in some cases, BDDs are competitive with d-DNNF thanks to the use of highly optimized packages.

\subsection{MAP Results}
For these experiments we ran PITA and ProbLog2.1 with the command \texttt{problog-cli.py map} \textit{program.pl}.
As MAP assignments of ground atoms must be explicitly queried, PITA requires the specification of the keyword \verb|map_query|  in front of the desired clauses.
Analogously, ProbLog2.1 uses the keyword \verb|query| that however can only be applied to probabilistic facts; so, for the datasets containing  clauses with multiple probabilistic heads, a syntactical transformation was applied before specifying the \verb|query| ground atoms.

For  \textit{gh} we used the program of size 11, containing 19 probabilistic clauses, and queried the 10\%, 20\%,...,90\% of them. For \textit{gnb} we used the program of size 10, containing 46 probabilistic clauses, and for \textit{blood} the  program of size 1, having 31 probabilistic clauses. For \textit{probabilistic graphs}, for each of the 10 edge configurations for each graph size $N$, we queried 20\%, 50\% and 80\% of the clauses:  the    50-node graphs contain 96 probabilistic  \verb|edge| facts, the   100-node graphs contain 196  \verb|edge| facts,  until the  500-node graphs which contain 996  \verb|edge| facts. We could not use the maximum size LPADs for \textit{gh}, \textit{gnb} and \textit{blood} due to memory errors or time-outs ($>24h$), hence we chose   a program size for which we could get  results in a reasonable time.
Evidence is the one specified in Section \ref{mpe_res}.
Inference times are compared in Table \ref{tab:map1} for \textit{gh, gnb, blood} and in Table~\ref{tab:map2} for  \textit{probabilistic graphs} with $N=50$; for $N \geq 100$ only PITA gave a result (almost always $<1min$, maximum time $=10min$ with $N=500$), while ProbLog2.1 always gave memory error or an error from the program. As expected, MAP inference takes more time, especially on \textit{gh} and \textit{blood}; PITA performs better than ProbLog on all datasets except \textit{blood}, indicating that BDD reordering is advantageous with respect to the use of ADDs.

\begin{table}[p]
	\scriptsize
	\caption{\small MAP inference time (s) comparison on the \textit{growing head, growing negated body, blood} datasets. In bold the best results. ``t-o" means time-out ($>$24hours),``me" means memory error.}
	\label{tab:map1}
	\begin{tabular*}{.3\textwidth}{lcc} \hline
		\multicolumn{3}{c}{\textit{Growing head}}\\\hline
		&ProbLog2.1      & PITA     \\ \hline\hline
		10\%    & 402.547     &\textbf{1.802} \\ \hline
		20\%   &860.220   &\textbf{0.547}\\ \hline
		30\%   &394.450    &\textbf{0.711}\\ \hline
		40\%   &2267.646      &\textbf{0.913}  \\ \hline
		50\%   &2436.738  &\textbf{0.949} \\ \hline
		60\%   &6420.507    &\textbf{2.315}\\ \hline
		70\%   & t-o   &\textbf{10.805}    \\ \hline
		80\%   & me  &\textbf{119.071}\\ \hline
		90\%   &  me  &\textbf{2520.562}\\ \hline
	\end{tabular*}
	\quad
	\begin{tabular*}{.3\textwidth}{lcc} \hline
		\multicolumn{3}{c}{\textit{Growing negated body}}\\\hline
		&ProbLog2.1      & PITA     \\ \hline\hline
		10\%  &\textbf{0.332}	&0.486         \\ \hline
		20\% &0.825	&\textbf{0.544}     \\ \hline
		30\%   &18.429	 &\textbf{0.559}    \\ \hline
		40\%   &477.893	&\textbf{0.648}        \\ \hline
		50\%    &30687.162 	&\textbf{0.797}  \\ \hline
		60\%   & t-o  &\textbf{1.161} \\ \hline
		70\%   & me    &\textbf{1.510}  \\ \hline
		80\%   & me &\textbf{1.388} \\ \hline
		90\%   & me  &\textbf{0.918} \\ \hline
	\end{tabular*}
	\quad
	\begin{tabular*}{.3\textwidth}{lcc} \hline
		\multicolumn{3}{c}{\textit{Blood}}\\\hline
		&ProbLog2.1      & PITA     \\ \hline\hline
		10\%      &\textbf{1.105}	&1.778     \\ \hline
		20\%      &\textbf{8.663}     &3576.321 \\ \hline
		30\%     &\textbf{836.331}	  & t-o  \\ \hline
		40\%      &\textbf{79957.043}	& t-o     \\ \hline
		50\%      & me    & me\\ \hline
		60\%      & me & me\\ \hline
		70\%      & me    & me\\ \hline
		80\%      & me	& me \\ \hline
		90\%      & me  & me\\ \hline
	\end{tabular*}
\end{table}

\begin{table}
	\scriptsize
	\caption{\small MAP inference time (s) comparison on the \textit{probabilistic graphs} dataset with $N=50$. In bold the best results. ``t-o" means time-out ($>$24hours),``me" means memory error, ``pe" means program error. ``PL" stands for ProbLog2.1.}
	\label{tab:map2}
	\subfloat{
		\begin{tabular}{lcc|cc|cc|cc|cc} \hline
			\multicolumn{3}{c}{\textit{Graph 1}} & \multicolumn{2}{c}{\textit{Graph 2}} & \multicolumn{2}{c}{\textit{Graph 3}} & \multicolumn{2}{c}{\textit{Graph 4}} & \multicolumn{2}{c}{\textit{Graph 5}}                                                                               \\\hline
			& PL                                   & PITA                                 & PL                                   & PITA                                 & PL       & PITA           & PL       & PITA           & PL & PITA           \\ \hline
			20\%                                 & pe                                   & \textbf{0.567}                       & pe                                   & \textbf{1.907}                       & 3789.323 & \textbf{2.075} & 4997.582 & \textbf{1.294} & pe & \textbf{1.721} \\ \hline
			50\%                                 & me                                   & \textbf{0.598}                       & me                                   & \textbf{0.702}                       & me       & \textbf{0.608} & me       & \textbf{0.584} & me & \textbf{0.628} \\ \hline
			80\%                                 & me                                   & \textbf{0.619}                       & me                                   & \textbf{0.686}                       & me       & \textbf{0.623} & me       & \textbf{0.598} & me & \textbf{0.632} \\ \hline
		\end{tabular}
	}
	\\ \subfloat{
		\begin{tabular}{lcc|cc|cc|cc|cc} \hline
			\multicolumn{3}{c}{\textit{Graph 6}} & \multicolumn{2}{c}{\textit{Graph 7}} & \multicolumn{2}{c}{\textit{Graph 8}} & \multicolumn{2}{c}{\textit{Graph 9}} & \multicolumn{2}{c}{\textit{Graph 10}}                                                                               \\\hline
			& PL                                   & PITA                                 & PL                                   & PITA                                  & PL       & PITA           & PL & PITA           & PL       & PITA           \\ \hline
			20\%                                 & 4596.1                               & \textbf{2.393}                       & 2812.411                             & \textbf{0.951}                        & 3187.299 & \textbf{1.713} & pe & \textbf{1.712} & 2955.692 & \textbf{1.733} \\ \hline
			50\%                                 & me                                   & \textbf{0.637}                       & me                                   & \textbf{9.174}                        & me       & \textbf{0.617} & me & \textbf{0.661} & me       & \textbf{0.588} \\ \hline
			80\%                                 & me                                   & \textbf{0.547}                       & me                                   & \textbf{1.140}                        & me       & \textbf{3.066} & me & \textbf{0.559} & me       & \textbf{0.598} \\ \hline
		\end{tabular}
	}
\end{table}

\section{Conclusions}
\label{sec:conclusions}

In this paper,
we presented an algorithm to solve
the Maximum-A-Posteriori (MAP)
and the Most-Probable-Explanation (MPE)
problems on Logic Programs with Annotated Disjunctions.
We integrated the algorithm into the PITA solver,
which is available as a SWI-Prolog package
and in the cplint on SWISH web application \cite{AlbCotRigZes16-AIIA-IC,cplint_ia_2017}
at \url{http://cplint.eu}.
We experimentally compared the algorithm 
with the ProbLog version (2.1) that supports annotated disjunctions and can perform the MAP and MPE tasks.
The results on several synthetic datasets show that
PITA performs better than ProbLog in many cases.  From our experimentation, we can conclude that since d-DNNF are theoretically better than BDD, one should first try ProbLog. In case the running time is high, however, using BDDs with PITA is an option to be considered because we demonstrated that in some cases the performance may be better.

In the future we plan to investigate the algorithm for finding Viterbi proofs \cite{DBLP:conf/ilp/ShterionovRVKMJ14}, i.e., partial truth value assignments (or partial possible worlds) such that for all full assignments extending the proof, the query holds.

\section*{Acknowledgments}

This work was partly supported by the ``National Group of Computing Science (GNCS-INDAM)''.

\bibliographystyle{acmtrans}

\begin{thebibliography}{}
	
	\bibitem[\protect\citeauthoryear{Alberti, Bellodi, Cota, Riguzzi, and
		Zese}{Alberti et~al\mbox{.}}{2017}]{cplint_ia_2017}
	{\sc Alberti, M.}, {\sc Bellodi, E.}, {\sc Cota, G.}, {\sc Riguzzi, F.}, {\sc
		and} {\sc Zese, R.} 2017.
	\newblock \texttt{cplint} on {SWISH}: Probabilistic logical inference with a
	web browser.
	\newblock {\em Intell. Artif.\/}~{\em 11,\/}~1, 47--64.
	
	\bibitem[\protect\citeauthoryear{Alberti, Cota, Riguzzi, and Zese}{Alberti
		et~al\mbox{.}}{2016}]{AlbCotRigZes16-AIIA-IC}
	{\sc Alberti, M.}, {\sc Cota, G.}, {\sc Riguzzi, F.}, {\sc and} {\sc Zese, R.}
	2016.
	\newblock Probabilistic logical inference on the web.
	\newblock In {\em AI*IA 2016: Advances in Artificial Intelligence, 21st
		Congress of the Italian Association for Artificial Intelligence, Pisa},
	{G.~Adorni}, {S.~Cagnoni}, {M.~Gori}, {and} {M.~Maratea}, Eds. Lecture Notes
	in Computer Science, vol. 10037. Springer International Publishing, 351--363.
	
	\bibitem[\protect\citeauthoryear{Barabasi and Albert}{Barabasi and
		Albert}{1999}]{Barabasi99}
	{\sc Barabasi, A.-L.} {\sc and} {\sc Albert, R.} 1999.
	\newblock Emergence of scaling in random networks.
	\newblock {\em Science\/}~{\em 286,\/}~5439, 509--512.
	
	\bibitem[\protect\citeauthoryear{Darwiche}{Darwiche}{2004}]{DBLP:conf/ecai/Darwiche04}
	{\sc Darwiche, A.} 2004.
	\newblock New advances in compiling {CNF} into decomposable negation normal
	form.
	\newblock In {\em 16th European Conference on Artificial Intelligence (ECAI
		20014)}, {R.~L. de~M{\'{a}}ntaras} {and} {L.~Saitta}, Eds. {IOS} Press,
	328--332.
	
	\bibitem[\protect\citeauthoryear{Darwiche and Marquis}{Darwiche and
		Marquis}{2002}]{DBLP:journals/jair/DarwicheM02}
	{\sc Darwiche, A.} {\sc and} {\sc Marquis, P.} 2002.
	\newblock A knowledge compilation map.
	\newblock {\em J. Artif. Intell. Res.\/}~{\em 17}, 229--264.
	
	\bibitem[\protect\citeauthoryear{De~Raedt, Demoen, Fierens, Gutmann, Janssens,
		Kimmig, Landwehr, Mantadelis, Meert, Rocha, Santos~Costa, Thon, and
		Vennekens}{De~Raedt et~al\mbox{.}}{2008}]{DeR-NIPS08}
	{\sc De~Raedt, L.}, {\sc Demoen, B.}, {\sc Fierens, D.}, {\sc Gutmann, B.},
	{\sc Janssens, G.}, {\sc Kimmig, A.}, {\sc Landwehr, N.}, {\sc Mantadelis,
		T.}, {\sc Meert, W.}, {\sc Rocha, R.}, {\sc Santos~Costa, V.}, {\sc Thon,
		I.}, {\sc and} {\sc Vennekens, J.} 2008.
	\newblock Towards digesting the alphabet-soup of statistical relational
	learning.
	\newblock In {\em {NIPS 2008} Workshop on Probabilistic Programming}.
	
	\bibitem[\protect\citeauthoryear{{De Raedt}, Frasconi, Kersting, and
		Muggleton}{{De Raedt} et~al\mbox{.}}{2008}]{DBLP:conf/ilp/2008p}
	{\sc {De Raedt}, L.}, {\sc Frasconi, P.}, {\sc Kersting, K.}, {\sc and} {\sc
		Muggleton, S.}, Eds. 2008.
	\newblock {\em Probabilistic Inductive Logic Programming}. LNCS, vol. 4911.
	Springer.
	
	\bibitem[\protect\citeauthoryear{{De Raedt}, Kimmig, and Toivonen}{{De Raedt}
		et~al\mbox{.}}{2007}]{DBLP:conf/ijcai/RaedtKT07}
	{\sc {De Raedt}, L.}, {\sc Kimmig, A.}, {\sc and} {\sc Toivonen, H.} 2007.
	\newblock {ProbLog}: A probabilistic {Prolog} and its application in link
	discovery.
	\newblock In {\em 20th International Joint Conference on Artificial
		Intelligence (IJCAI 2007)}, {M.~M. Veloso}, Ed. Vol.~7. AAAI Press/IJCAI,
	2462--2467.
	
	\bibitem[\protect\citeauthoryear{Fierens, {Van den Broeck}, Renkens,
		Shterionov, Gutmann, Thon, Janssens, and {De Raedt}}{Fierens
		et~al\mbox{.}}{2015}]{DBLP:journals/corr/abs-1304-6810}
	{\sc Fierens, D.}, {\sc {Van den Broeck}, G.}, {\sc Renkens, J.}, {\sc
		Shterionov, D.~S.}, {\sc Gutmann, B.}, {\sc Thon, I.}, {\sc Janssens, G.},
	{\sc and} {\sc {De Raedt}, L.} 2015.
	\newblock Inference and learning in probabilistic logic programs using weighted
	{Boolean} formulas.
	\newblock {\em Theor. Pract. Log. Prog.\/}~{\em 15,\/}~3, 358--401.
	
	\bibitem[\protect\citeauthoryear{Jiang, Babar, Ciardo, Miner, and Smith}{Jiang
		et~al\mbox{.}}{2017}]{jiang2017variable}
	{\sc Jiang, C.}, {\sc Babar, J.}, {\sc Ciardo, G.}, {\sc Miner, A.~S.}, {\sc
		and} {\sc Smith, B.} 2017.
	\newblock Variable reordering in binary decision diagrams.
	\newblock In {\em 26th International Workshop on Logic and Synthesis}. 1--8.
	
	\bibitem[\protect\citeauthoryear{Poole}{Poole}{1997}]{Poo97-ArtInt-IJ}
	{\sc Poole, D.} 1997.
	\newblock The {I}ndependent {C}hoice {L}ogic for modelling multiple agents
	under uncertainty.
	\newblock {\em Artif. Intell.\/}~{\em 94}, 7--56.
	
	\bibitem[\protect\citeauthoryear{Poole}{Poole}{2000}]{DBLP:journals/jlp/Poole00}
	{\sc Poole, D.} 2000.
	\newblock Abducing through negation as failure: Stable models within the
	independent choice logic.
	\newblock {\em J. Logic Program.\/}~{\em 44,\/}~1-3, 5--35.
	
	\bibitem[\protect\citeauthoryear{Raedt, Kimmig, and Toivonen}{Raedt
		et~al\mbox{.}}{2007}]{raedt07problog}
	{\sc Raedt, L.~D.}, {\sc Kimmig, A.}, {\sc and} {\sc Toivonen, H.} 2007.
	\newblock Problog: A probabilistic prolog and its application in link
	discovery.
	\newblock In {\em IJCAI}, {M.~M. Veloso}, Ed. 2462--2467.
	
	\bibitem[\protect\citeauthoryear{Riguzzi}{Riguzzi}{2016}]{Rig16-IJAR-IJ}
	{\sc Riguzzi, F.} 2016.
	\newblock The distribution semantics for normal programs with function symbols.
	\newblock {\em Int. J. Approx. Reason.\/}~{\em 77}, 1--19.
	
	\bibitem[\protect\citeauthoryear{Riguzzi}{Riguzzi}{2018}]{Rig18-BK}
	{\sc Riguzzi, F.} 2018.
	\newblock {\em Foundations of Probabilistic Logic Programming}.
	\newblock River Publishers, Gistrup,Denmark.
	
	\bibitem[\protect\citeauthoryear{Riguzzi and Swift}{Riguzzi and
		Swift}{2010}]{RigSwi10-ICLP10-IC}
	{\sc Riguzzi, F.} {\sc and} {\sc Swift, T.} 2010.
	\newblock Tabling and answer subsumption for reasoning on logic programs with
	annotated disjunctions.
	\newblock In {\em Technical Communications of the 26th International Conference
		on Logic Programming ({ICLP} 2010)}. LIPIcs, vol.~7. Schloss Dagstuhl -
	Leibniz-Zentrum fuer Informatik, 162--171.
	
	\bibitem[\protect\citeauthoryear{Riguzzi and Swift}{Riguzzi and
		Swift}{2011}]{RigSwi11-ICLP11-IJ}
	{\sc Riguzzi, F.} {\sc and} {\sc Swift, T.} 2011.
	\newblock The {PITA} system: Tabling and answer subsumption for reasoning under
	uncertainty.
	\newblock {\em Theor. Pract. Log. Prog.\/}~{\em 11,\/}~4--5, 433--449.
	
	\bibitem[\protect\citeauthoryear{Riguzzi and Swift}{Riguzzi and
		Swift}{2013}]{RigSwi13-TPLP-IJ}
	{\sc Riguzzi, F.} {\sc and} {\sc Swift, T.} 2013.
	\newblock Well\--definedness and efficient inference for probabilistic logic
	programming under the distribution semantics.
	\newblock {\em Theor. Pract. Log. Prog.\/}~{\em 13,\/}~2, 279--302.
	
	\bibitem[\protect\citeauthoryear{Sato}{Sato}{1995}]{DBLP:conf/iclp/Sato95}
	{\sc Sato, T.} 1995.
	\newblock A statistical learning method for logic programs with distribution
	semantics.
	\newblock In {\em Logic Programming, Proceedings of the Twelfth International
		Conference on Logic Programming, Tokyo, Japan, June 13-16, 1995},
	{L.~Sterling}, Ed. MIT Press, 715--729.
	
	\bibitem[\protect\citeauthoryear{Shterionov, Renkens, Vlasselaer, Kimmig,
		Meert, and Janssens}{Shterionov
		et~al\mbox{.}}{2015}]{DBLP:conf/ilp/ShterionovRVKMJ14}
	{\sc Shterionov, D.~S.}, {\sc Renkens, J.}, {\sc Vlasselaer, J.}, {\sc Kimmig,
		A.}, {\sc Meert, W.}, {\sc and} {\sc Janssens, G.} 2015.
	\newblock The most probable explanation for probabilistic logic programs with
	annotated disjunctions.
	\newblock In {\em 24th International Conference on Inductive Logic Programming
		(ILP 2014)}, {J.~Davis} {and} {J.~Ramon}, Eds. Lecture Notes in Computer
	Science, vol. 9046. Springer, Berlin, Heidelberg, 139--153.
	
	\bibitem[\protect\citeauthoryear{Somenzi}{Somenzi}{2001}]{DBLP:journals/sttt/Somenzi01}
	{\sc Somenzi, F.} 2001.
	\newblock Efficient manipulation of decision diagrams.
	\newblock {\em Int. J. Softw. Tools Technol. Transf.\/}~{\em 3,\/}~2, 171--181.
	
	\bibitem[\protect\citeauthoryear{Valiant}{Valiant}{1979}]{valiant1979complexity}
	{\sc Valiant, L.~G.} 1979.
	\newblock The complexity of enumeration and reliability problems.
	\newblock {\em SIAM J. Comput.\/}~{\em 8,\/}~3, 410--421.
	
	\bibitem[\protect\citeauthoryear{Van~den Broeck, Thon, van Otterlo, and
		De~Raedt}{Van~den Broeck et~al\mbox{.}}{2010}]{DBLP:conf/aaai/BroeckTOR10}
	{\sc Van~den Broeck, G.}, {\sc Thon, I.}, {\sc van Otterlo, M.}, {\sc and} {\sc
		De~Raedt, L.} 2010.
	\newblock {DTProbLog}: A decision-theoretic probabilistic {Prolog}.
	\newblock In {\em Proceedings of the Twenty-Fourth AAAI Conference on
		Artificial Intelligence}, {M.~Fox} {and} {D.~Poole}, Eds. AAAI Press,
	1217--1222.
	
	\bibitem[\protect\citeauthoryear{Vennekens, Verbaeten, and
		Bruynooghe}{Vennekens et~al\mbox{.}}{2004a}]{VenVer04-ICLP04-IC}
	{\sc Vennekens, J.}, {\sc Verbaeten, S.}, {\sc and} {\sc Bruynooghe, M.} 2004a.
	\newblock Logic programs with annotated disjunctions.
	\newblock In {\em 24th International Conference on Logic Programming (ICLP
		2004)}, {B.~Demoen} {and} {V.~Lifschitz}, Eds. Lecture Notes in Computer
	Science, vol. 3131. Springer, 431--445.
	
	\bibitem[\protect\citeauthoryear{Vennekens, Verbaeten, and
		Bruynooghe}{Vennekens et~al\mbox{.}}{2004b}]{VenVer04-ICLP04-IC_iclp}
	{\sc Vennekens, J.}, {\sc Verbaeten, S.}, {\sc and} {\sc Bruynooghe, M.} 2004b.
	\newblock Logic programs with annotated disjunctions.
	\newblock In {\em 24th International Conference on Logic Programming (ICLP
		2004)}, {B.~Demoen} {and} {V.~Lifschitz}, Eds. Lecture Notes in Computer
	Science, vol. 3131. Springer, Berlin, 195--209.
	
\end{thebibliography}


\end{document}